\def\year{2020}\relax
\documentclass[letterpaper]{article} 
\usepackage{aaai20}  
\usepackage{times}  
\usepackage{helvet} 
\usepackage{courier}  
\usepackage[hyphens]{url}  
\usepackage{graphicx} 
\usepackage{graphicx}  
\usepackage{array}
\newcolumntype{P}[1]{>{\centering\arraybackslash}p{#1}}

\usepackage{amsmath}
\usepackage{amsfonts}
\usepackage{graphicx}
\usepackage{booktabs}
\usepackage{enumerate}
\usepackage{multirow}

\usepackage{arydshln}
\usepackage[english]{babel}
\usepackage[autostyle, english=american]{csquotes}
\MakeOuterQuote{"}
\usepackage{subcaption}
\usepackage{soul}
\usepackage{mwe}

\title{\textsc{AvgOut}: A Simple Output-Probability Measure to \\ Eliminate Dull Responses}
\author{
\Large \textbf{Tong Niu, Mohit Bansal}\\ 
UNC Chapel Hill\\ 
$\{$tongn, mbansal$\}$@cs.unc.edu 
}

\begin{document}

\maketitle

\begin{abstract}
    Many sequence-to-sequence dialogue models tend to generate safe, uninformative responses. There have been various useful efforts on trying to eliminate them. However, these approaches either improve decoding algorithms during inference, rely on hand-crafted features, or employ complex models. In our work, we build dialogue models that are dynamically aware of what utterances or tokens are dull without any feature-engineering. Specifically, we start with a simple yet effective automatic metric, \textsc{AvgOut}, which calculates the average output probability distribution of all time steps on the decoder side during training. This metric directly estimates which tokens are more likely to be generated, thus making it a faithful evaluation of the model diversity (i.e., for diverse models, the token probabilities should be more evenly distributed rather than peaked at a few dull tokens). We then leverage this novel metric to propose three models that promote diversity without losing relevance. The first model, \textsc{MinAvgOut}, directly maximizes the diversity score through the output distributions of each batch; the second model, Label Fine-Tuning (\textsc{LFT}), prepends to the source sequence a label continuously scaled by the diversity score to control the diversity level; the third model, \textsc{RL}, adopts Reinforcement Learning and treats the diversity score as a reward signal. Moreover, we experiment with a hybrid model by combining the loss terms of \textsc{MinAvgOut} and \textsc{RL}. All four models outperform their base LSTM-RNN model on both diversity and relevance by a large margin, and are comparable to or better than competitive baselines (also verified via human evaluation). Moreover, our approaches are orthogonal to the base model, making them applicable as an add-on to other emerging better dialogue models in the future.
\end{abstract}

\section{Introduction}
\label{sec:introduction}
Many modern dialogue generation models use a sequence-to-sequence architecture as their backbone~\cite{sordoni2015neural}, following its success when applied to Machine Translation (\textsc{MT})~\cite{Bahdanau2015}. However, dialogue tasks also have a requirement different from that of \textsc{MT}: the response not only has to be "correct" (coherent and relevant), but also needs to be diverse and informative. However, seq2seq has been reported by many previous works to have low corpus-level diversity~\cite{li2015diversity,serban2016building,sordoni2015neural,vinyals2015neural}, as it tends to generate safe, terse, and uninformative responses, such as "\textit{I don't know.}".
These responses unnecessarily make a dialogue system much less interactive than it should be.

To increase the diversity of dialogue responses, 
the first step is to faithfully evaluate how diverse a response is. There are metrics used by previous work that are correlated to diversity, but not strongly, such as ratio of distinct tokens~\cite{li2015diversity} and response length~\cite{baheti2018generating}. However, a response can be long but extremely boring in meaning, such as \textit{"I am sure that I don't know about it."}, or short but interesting (i.e., contains a lot of information), such as \textit{"Dad was mean."}. Only investigating discrete token output by the model is also not ideal, because these tokens are only a single realization of the model's output probability distribution at each time step, which unavoidably loses valuable information indicated by the whole distribution.~\citeauthor{li2016deep} (\citeyear{li2016deep}) manually collect a shortlist of dull responses, and during training discourage the model from producing such utterances. However, an important drawback of hand-crafted rules is that the set of dull tokens or utterances is static, while in fact it usually evolves during training: when the current dull tokens are eliminated, another set of them might reveal themselves. 

In our work,\footnote{We will release all our code and model outputs.} we begin with a simple yet effective approach to measure how diverse a response is. This metric, which we name "Average Output Probability Distribution", or \textsc{AvgOut}, draws information directly from the training-in-session model itself. We calculate it by keeping track of the exponential average of all output probability distributions on the decoder side during training. This metric dynamically measures which tokens the model is biased toward without any hand-crafted rules, thus making it a faithful evaluation of the model diversity (i.e., for diverse models, the token probabilities should be more evenly distributed rather than peaked at a few dull tokens). In addition, since \textsc{AvgOut} is a \textit{one-dimensional categorical distribution} rather than a dimensionless numerical value like entropy, it naturally carries and conveys more information about model diversity.

We then propose three models that leverage our novel metric to promote diversity in dialogue generation. The first \textsc{MinAvgOut} model minimizes the dot product of current batch \textsc{AvgOut} and the exponential average \textsc{AvgOut} across batches, which encourages low-frequency tokens to be generated. The second \textsc{LFT} model uses a labeled transduction method and scales a "diversity label" by the diversity score of the ground-truth target sequence during training, while during testing can generate responses of different levels of diversity by tweaking the intended diversity score. The third \textsc{RL} model leverages reinforcement learning, where our novel metric is applied to discrete tokens and serve as a reward signal. In addition, since \textsc{MinAvgOut} regularizes directly on the continuous distribution while \textsc{RL} calculates its reward based on discrete sampled tokens, we simply add up the loss terms of the two models, creating an even stronger hybrid model.

We first employ diverse automatic metrics, including Distinct-1 and -2 from previous work~\cite{li2015diversity} and our novel metric~\textsc{Diveristy-}\textit{i}\textsc{AUC} (which calculates one minus the sum of normalized frequencies of the most frequent tokens produced by the model), plus activity/entity F1s, 
to evaluate the diversity and relevance of the generated responses. We then conduct human evaluations to verify that these models not only outperform their base model \textsc{LSTM} by a large margin, but are also comparable to or better than an advanced decoding algorithm \textsc{MMI}~\cite{li2015diversity} and a very competitive model \textsc{VHRED}~\cite{serban2017hierarchical} on the Ubuntu dataset.

\section{\textsc{AvgOut} as an Effective Diversity Metric}
By only keeping a static shortlist of boring responses or tokens, one basically assumes that we humans should decide which tokens are dull. However, we argue that we should instead look from the model's perspective to identify dull tokens, because even if the model outputs a word that we consider rare, including it in too many responses is still considered a dull behavior. Motivated by this thought experiment, we propose a novel metric, Average Output Probability Distribution (\textsc{AvgOut}), that dynamically keeps track of which tokens the model is biased toward. To calculate this, during training, we average out all the output probability distributions for each time step of the decoder for the whole mini-batch. The resulting vector $D'$ will reflect each token's probability of being generated from the model's perspective. Note that we do not use discrete ground-truth tokens to evaluate the model's bias, because there is a fine distinction between the two: a statistics of frequency on ground-truth tokens is an evaluation of \textit{the corpus's bias}, while \textsc{AvgOut} is an evaluation of \textit{what bias the model has learned} because by generating dull responses more frequently than the training corpus has, it is the model itself that we should adjust.
Also note that the reason we take the average is that a single output distribution will largely depend on the context and the previous target tokens (which are fed as inputs to the decoder during training), but on average the distribution should be a faithful evaluation on which words are more likely to be generated from model's perspective.

To avoid batches that have \textsc{AvgOut} significantly different from those of other batches, which would lead the model astray, we keep the exponential average of this metric across batches to make it less biased toward any specific batch.
Let it be $D$. After training on a mini-batch and obtain $D'$, we update $D$ like the following:
\begin{align*}
    D \leftarrow \gamma D' + (1 - \gamma) D
\end{align*}
where $\gamma$ is $0.01$ in our experiments.

Another consideration of \textsc{AvgOut} is that theoretically we can have two choices. The first is to use the output distributions when we are teacher-forcing (i.e., only feeding ground-truth tokens); the other is to let the model use its own predictions during greedy/beam-search decoding or sampling. We reason that the former is a much better estimation of the model's bias, because the latter will result in a cascading enlargement of the model bias due to the auto-regressive nature of LSTM-RNN models (i.e., the tokens fed to the decoder are themselves also polluted by the model's bias). Our early experimental results also agreed with the above reasoning. 

Although we try to come up with the most faithful evaluation of how diverse a response is, our approach certainly has its drawbacks too. For example, using very frequent words but less frequent combinations of them may result in a good response which will be penalized by our metric. A natural solution to this is to also use bigram and trigram diversities and take a linear combination of them, which on a high-level is similar to \textsc{BLEU}~\cite{Papineni:2002:BMA:1073083.1073135}. However, considering even bigram distribution takes up $O(|V|^2)$ space and calculation time, hence we did not try it due to limited resources. However, as will be presented in Section~\ref{sec:results and analysis}, regularizing unigram distributions can already greatly help on higher-gram diversities, while also improving relevance.

\section{Three Models to Leverage \textsc{AvgOut}}
\label{sec:model}

AvgOut can play at least three roles. First, it can be used to directly supervise output distribution during training; second, it can be used as a prior in labeled sequence transduction methods to control diversity of the generated response; and third, it can be used as a reward signal for Reinforcement Learning to encourage diverse sampled responses.\footnote{It can also be used to re-rank beams during inference, but we are much more interested in making the underlying model itself more diverse.} In this section, we begin with a base vanilla seq2seq model, and next present our three models to diversify responses based on \textsc{AvgOut}.

Our base model \textsc{LSTM} is identical to that proposed by~\citeauthor{Bahdanau2015} (\citeyear{Bahdanau2015}), which consists of a single-layer bi-directional LSTM-RNN~\cite{hochreiter1997long} encoder and a single-layer LSTM-RNN decoder with additive attention.

\begin{figure}[t]
\centering
\includegraphics[width=0.45\textwidth]{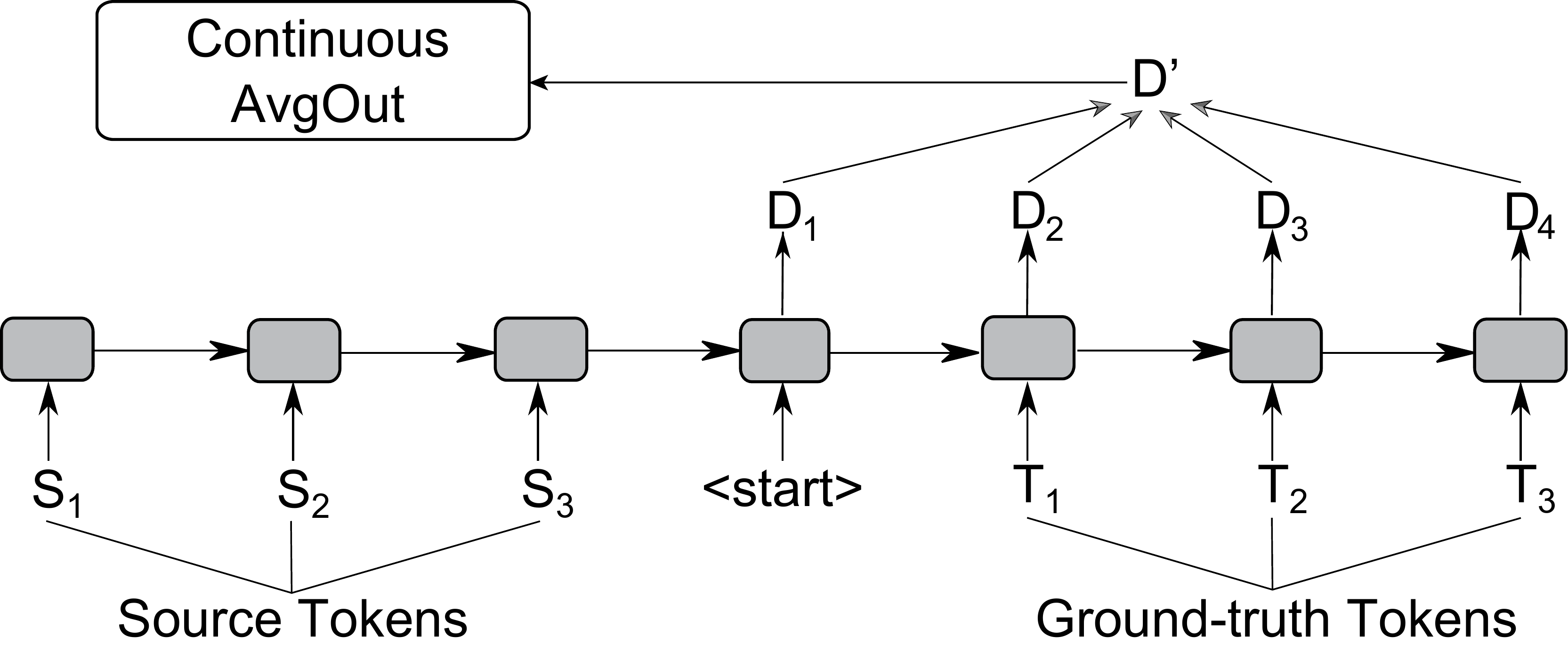}
\caption{MinAvgOut model: use the dot product of average output distribution of the exponential average and the current batch to evaluate how diverse the current batch is.}
\label{fig:MinAvgOut}
\end{figure}

\subsection{Regularization by Minimizing \textsc{Continuous-AvgOut}}
\begin{figure}[ht!]
\centering
\includegraphics[width=0.43\textwidth]{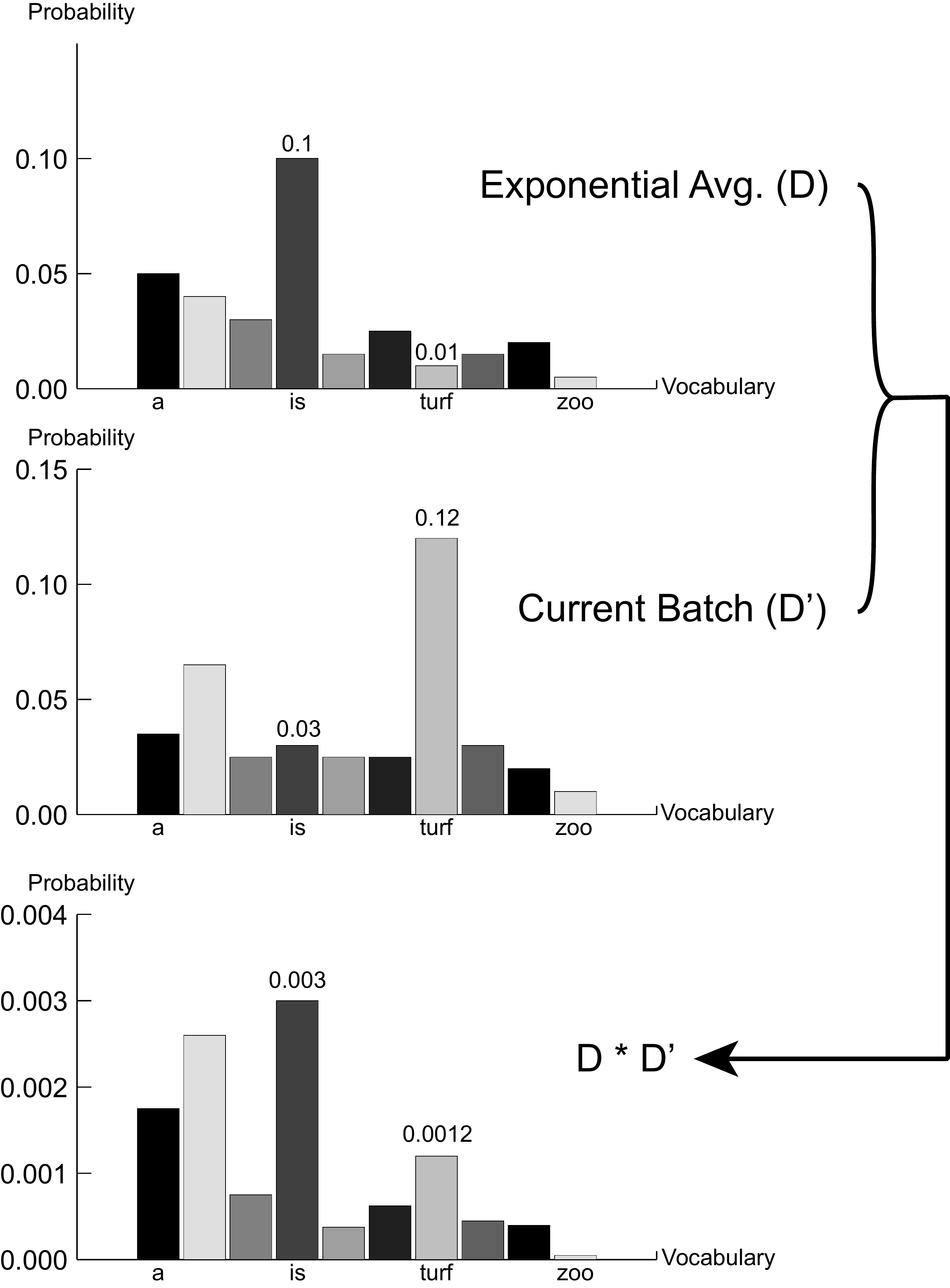}
\caption{An example of \textsc{AvgOut} applied to a single token, which readily generalizes to multiple tokens within a response. We calculate diversity score of a continuous distribution through dot product. We sum up the values in the last graph. Note that although "\textit{turf}" (the fourth word from the right in all three sub-figures) has higher probability in the current batch $D'$, it still contributes less than the word "\textit{is}" to the overall diversity measure when taking the dot product, due to its low probability in the exponential average distribution $D$ (i.e., lower weights). All probabilities are for illustration purpose and do not correspond to distributions from our models.
}
\label{fig:continuous}
\end{figure}

Our \textsc{MinAvgOut} model (Figure~\ref{fig:MinAvgOut}) directly integrates \textsc{AvgOut} into the loss function by summarizing it into a single numerical value named \textsc{Continuous-AvgOut}.
We do this by taking the dot-product of $D$ and $D'$ (Figure~\ref{fig:continuous}).
The intuition behind this simple calculation is that $D$ can also be viewed as a set of weights which add up to $1.0$, since it is a probability vector. By taking the dot product, we are actually calculating a weighted average of each probability in $D'$. To evaluate how diverse the model currently is, the duller tokens should obviously carry higher weights since they contribute more to the "dullness" of the whole utterance.\footnote{This linear combination is crucial, because if we naively add up all probabilities in $D'$ and take the average, the result will be a useless constant $1.0/|V|$, where $|V|$ is the vocabulary size. This naive approach certainly cannot capture the diversity of a model.} Assuming that $D$ is a column vector, the continuous diversity score is $B_c$, and the resulting extra loss term is $L_B$, the total loss $L$ is given by:
\begin{align*}
    B_c = 1 - D^T D' ; L_B = (-1) * \alpha B_c ; L = L_{ML} + L_B
\end{align*}
where $\alpha$ is a coefficient to balance the regularization loss with the maximum likelihood loss (a.k.a. teacher forcing loss) $L_{ML}$. This is important because the regularization term continues to discourage the model from generating the ground-truth token, which we need to balance by ML loss to reduce the impact (otherwise the model will be led astray). Note that since $D$ is a moving average which does not depend on the model parameters of the current mini-batch, only $D'$ will result in gradient flow during back-propagation, which is what we intend.

\subsection{Label-Fine-Tuning Model}
\begin{figure}[t]
\centering
\includegraphics[width=0.45\textwidth]{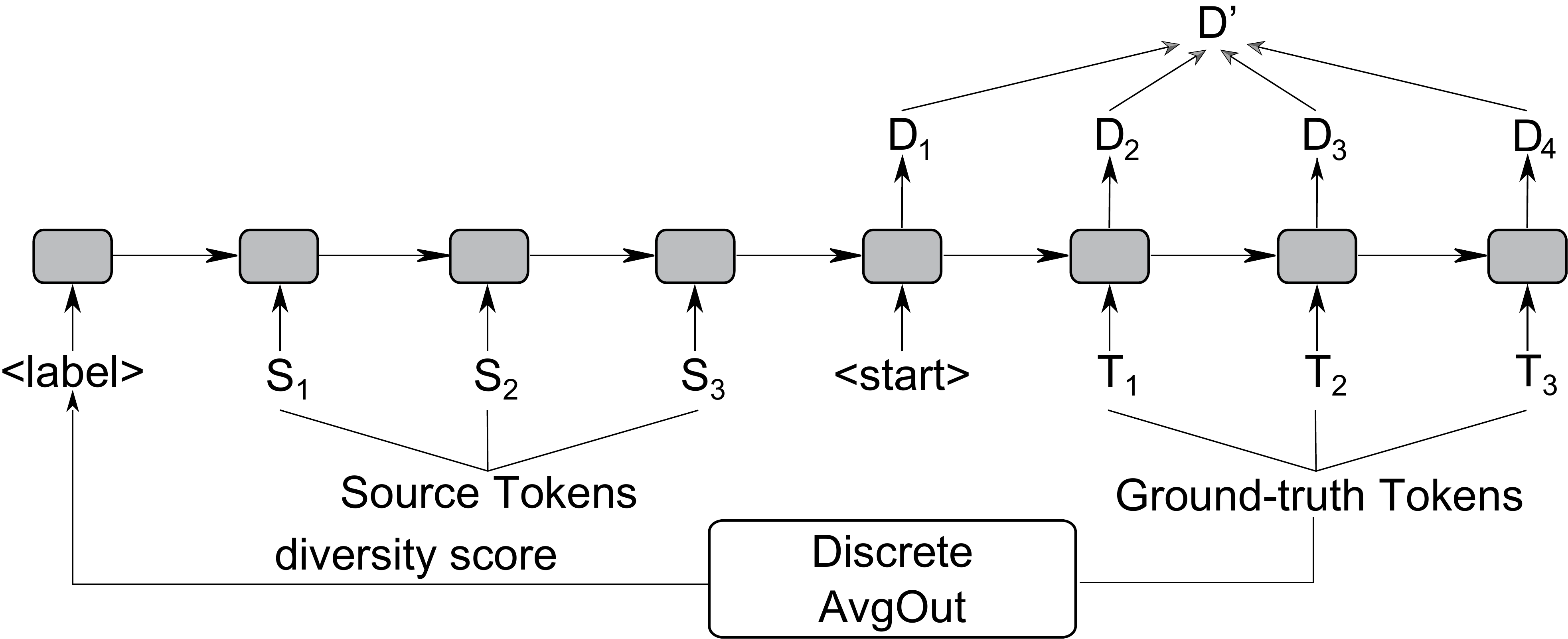}
\caption{\textsc{LFT} model: the diversity label is scaled by the diversity score of the ground-truth target during training.}
\label{fig:LFT}
\end{figure}

We also borrow the continuous version of the Label-Fine-Tuning (\textsc{LFT}) model from~\citeauthor{TACL1424} (\citeyear{TACL1424}), which is an extension of the discrete labeled sequence transduction methods~\cite{DBLP:journals/corr/KikuchiNSTO16}.
The LFT model leverages a continuous label to serve as a prior for generating the target sequence. This label corresponds to an embedding just like a normal token does, but can be scaled by a continuous value.
This model is applicable to our case because the diversity score of a response can also be viewed as a style, ranging from $0.0$ to $1.0$. Specifically, we add to the vocabulary a diversity label and scale its embedding vector with the intended diversity score of the target sequence. During training, this score is obtained by evaluating the diversity of the ground-truth target sequence (see Figure~\ref{fig:LFT}); during test time, we instead feed the model a diversity label scaled by a score of our choice (i.e., when we want the model to generate a more diverse response, we scale the label's embedding by a higher score, while to generate a duller response, we scale the embedding by a lower one).

\subsection{Reward-Based Reinforcement Learning}
\begin{figure}[t]
\centering
\includegraphics[width=0.46\textwidth]{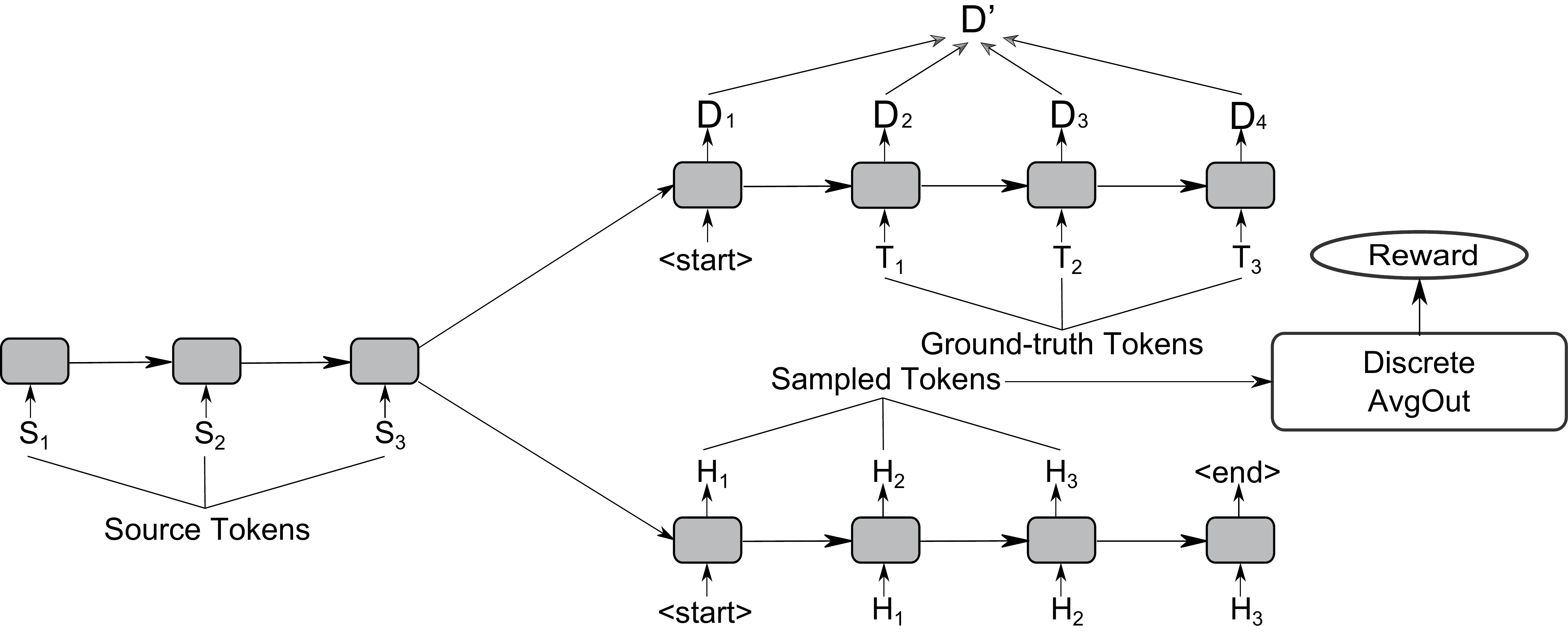}
\caption{RL model: the diversity score of the sampled response is fed back to the model as reward signal.}
\label{fig:RL}
\end{figure}

\begin{figure}[t]
\centering
\includegraphics[width=0.45\textwidth]{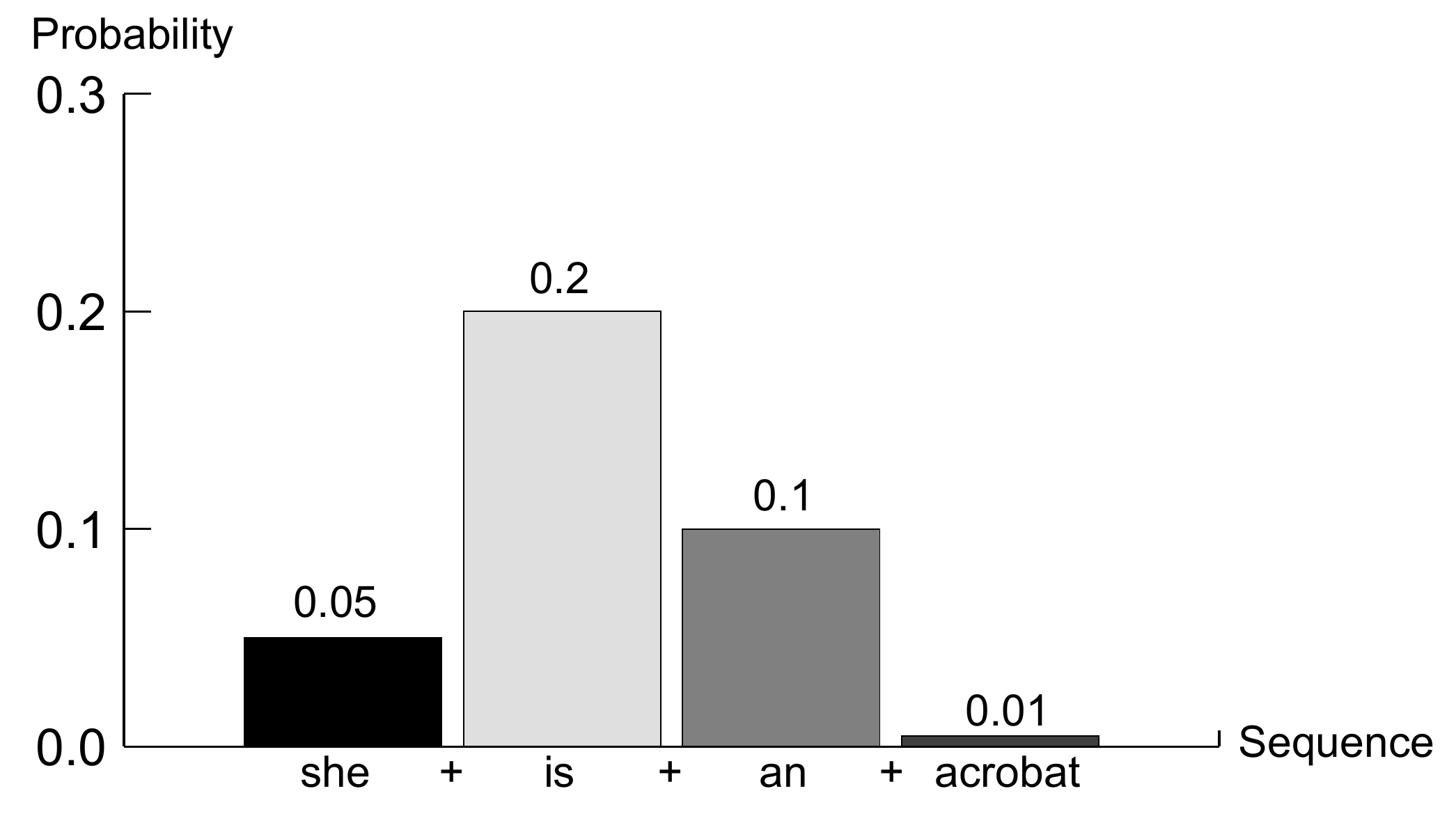}
\caption{Diversity score calculation for a discrete sequence: $B_d = 1.0 - (0.05 + 0.2 + 0.1 + 0.01) / 4$.}
\label{fig:discrete}
\end{figure}

We also explore a model (see Figure~\ref{fig:RL}) which regularizes on the discrete token level, because merely monitoring output probability distribution may ignore certain bad styles such as repetition (e.g. "\textit{I don't don't know.}"). We use \textsc{Discrete-AvgOut} to calculate the continuous diversity score of a discrete sequence. Let $\{G_1, G_2, ..., G_{N_G}\}$ be a sequence of $N_G$ tokens sampled by the model during training. Then from $D$, we extract the probabilities $\{P_1, P_2, ..., P_{N_G}\}$ corresponding to each generated token. The diversity score $B_{d}$ on these discrete tokens will be:
\begin{align*}
    B_d &= 1 - (P_1 + P_2 + ... + P_{N_G}) / N_{unique}
\end{align*}
where $N_{unique}$ is the number of unique tokens in the sampled sequence (see Figure~\ref{fig:discrete}).
Note that this division explicitly discourages the model from outputting repeated tokens, because when that happens, the nominator will stay the same, while the denominator will decrease, resulting in a lower diversity score. Also note that \textsc{MinAvgOut} can be complementary to \textsc{RL} since calculating diversity scores based on discrete tokens unavoidably loses valuable information from the output distribution before argmax is taken. In Section~\ref{sec:results and analysis}, we show with both automatic and human evaluations that this combination indeed achieves the best results among our models.
Following ~\citeauthor{DBLP:journals/corr/PaulusXS17} (\citeyear{DBLP:journals/corr/PaulusXS17}), our loss function consists of two terms. The first term is the Maximum Likelihood loss ($L_{\textsc{ml}}$); the other is the Reinforcement Learning loss ($L_{\textsc{rl}}$). The total loss $L$ is then:
\begin{align*}
\label{eq:rl}
	L &= L_{\textsc{ml}} + \beta \, L_{\textsc{rl}} \\
	L_{\textsc{ml}} &= - \sum^n_{t=1} \log p(y_t^* | y_1^*, ..., y_{t-1}^*, x) \\
	L_{\textsc{rl}} &= - \, (R - R_b) \, \sum^n_{t=1} \log p(y_t^s | y_1^s, ..., y_{t-1}^s, x)
\end{align*}
where $\beta$ is a hyperparameter indicating how much weight we want to assign to the \textsc{rl} part of the loss, $x$ is the source sequence, $\{y_t^*\}$ are the ground truth tokens and $\{y_t^s\}$ are the sampled tokens.
We use a policy gradient method~\cite{Sutton00policygradient} to calculate the \textsc{RL} loss. Specifically, we sample a response for each context $x$, and assign to it a reward $R$, which is equal to $B_d$ because we want to encourage the model to be diverse.
We also use a baseline $R_b$ that helps reduce variance during training~\cite{ranzato2015sequence}. In our case this baseline is again the exponential average of all $B_d$ in previous mini-batches.

\section{Experimental Setup}
\subsection{Dataset and Task}
We use the task-oriented Ubuntu Dialogue dataset~\cite{lowe2015ubuntu}, because it not only has F1 metrics to evaluate the relevance of responses, but the dialogues in them are also open-ended to allow enough space for diversity.
We also chose this dataset because previous work, e.g., HRED~\cite{serban2016building} and VHRED~\cite{serban2017hierarchical} both used Ubuntu to showcase their diversity-promotion models. Due to the popularity of this dataset, we were able to reproduce almost all models on this same dataset and have a meaningful comparison on their effectiveness of eliminating dullness.
As future work, we plan to apply our models to other datasets where diversity is desired.

\subsection{Automatic Evaluation}
To measure the relevance of the model responses, we follow~\citeauthor{serban2017multiresolution} (\citeyear{serban2017multiresolution}) and evaluate on F1's for both activities (technical verbs, e.g., "upload", "install")
and entities (technical nouns, e.g., "root", "internet"). The F1's are computed by mapping the ground-truth and model responses to their corresponding activity-entity representations~\cite{serban2017multiresolution}, who considered F1 to be "particularly suited for the goal-oriented Ubuntu Dialogue Corpus".
We did not evaluate on BLEU score~\cite{Papineni:2002:BMA:1073083.1073135} because~\cite{liu2016not} showed that BLEU does not correlate well with dialogue quality.~\citeauthor{lowe2017towards} (\citeyear{lowe2017towards}) also made similar observations on BLEU.
To evaluate diversity, we employ two evaluation metrics from previous work, namely \textsc{Distinct-1} and \textsc{Distinct-2}~\cite{li2015diversity}. These are the ratios between the number of unique tokens and all tokens for unigrams and bigrams, respectively. 
In addition, we propose a novel diversity graph and its corresponding metric, which we name \textsc{Diversity-$32$} and \textsc{Diversity-AUC}, respectively. We gather statistics of sentence, unigram, bigram and trigram, and sort their normalized frequencies from highest to lowest. Observing that all four graphs follow long-tail distributions, we only keep the highest $32$ frequencies and plot them.\footnote{One can also pick any reasonable number around $32$ without loss of generality. Also note that we do not take into account what specific token each frequency corresponds to, because for example, although ``\textit{plethora}'' is an unusual word, a model that outputs it all the time is still boring. Thus the frequencies are more important than what the tokens actually are.} We then calculate one minus the Area under Curve (\textsc{Diversity-AUC}) for each graph, which draws a high-level picture of how diverse a model is.

\begin{figure}[t]
    \centering
    \begin{subfigure}[b]{0.224\textwidth}
        \centering
        \includegraphics[width=\textwidth]{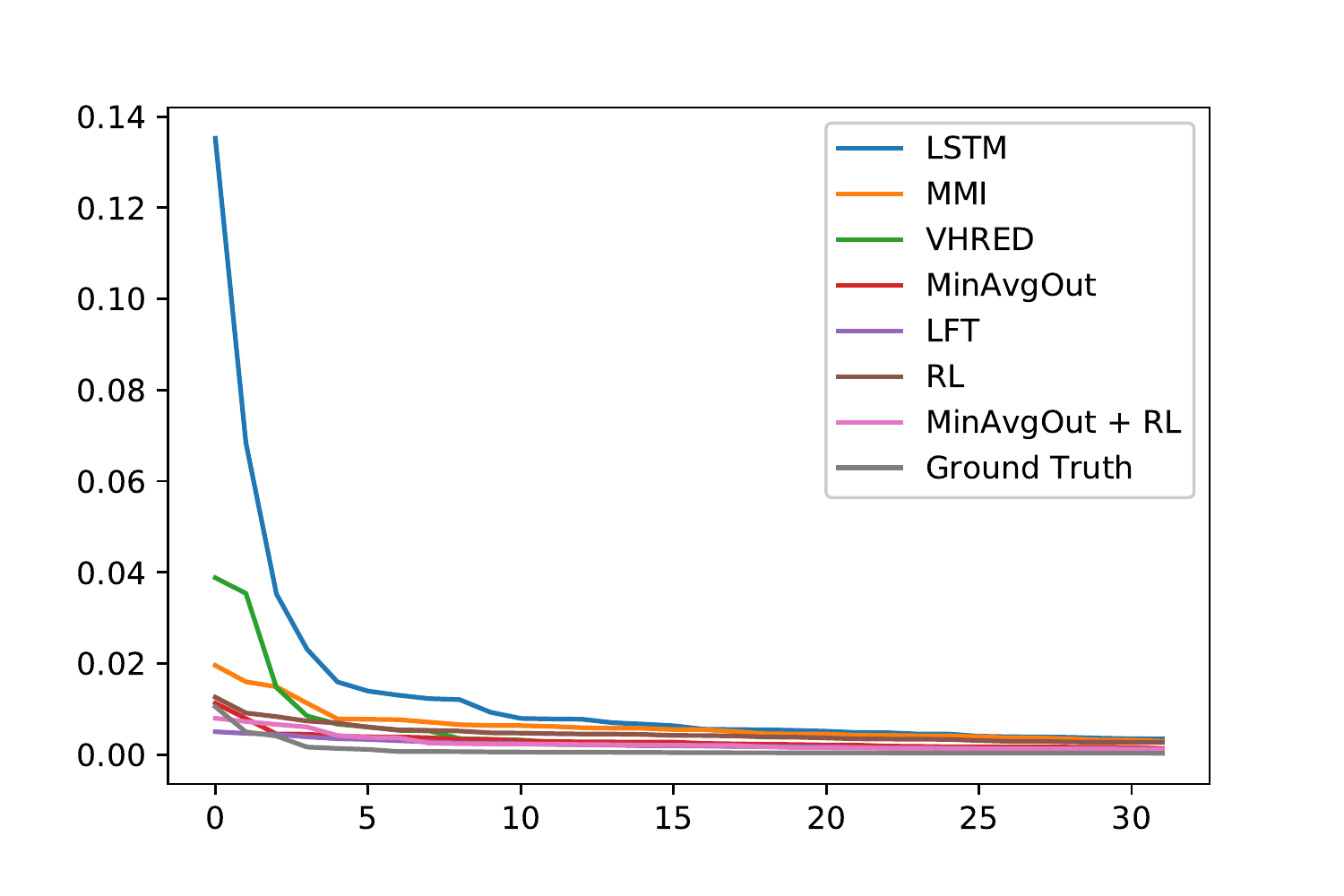}
        \caption[]%
        {{\small Sentence-Level}}
        \label{fig:diversity-sent}
    \end{subfigure}
    \hfill
    \begin{subfigure}[b]{0.224\textwidth}  
        \centering 
        \includegraphics[width=\textwidth]{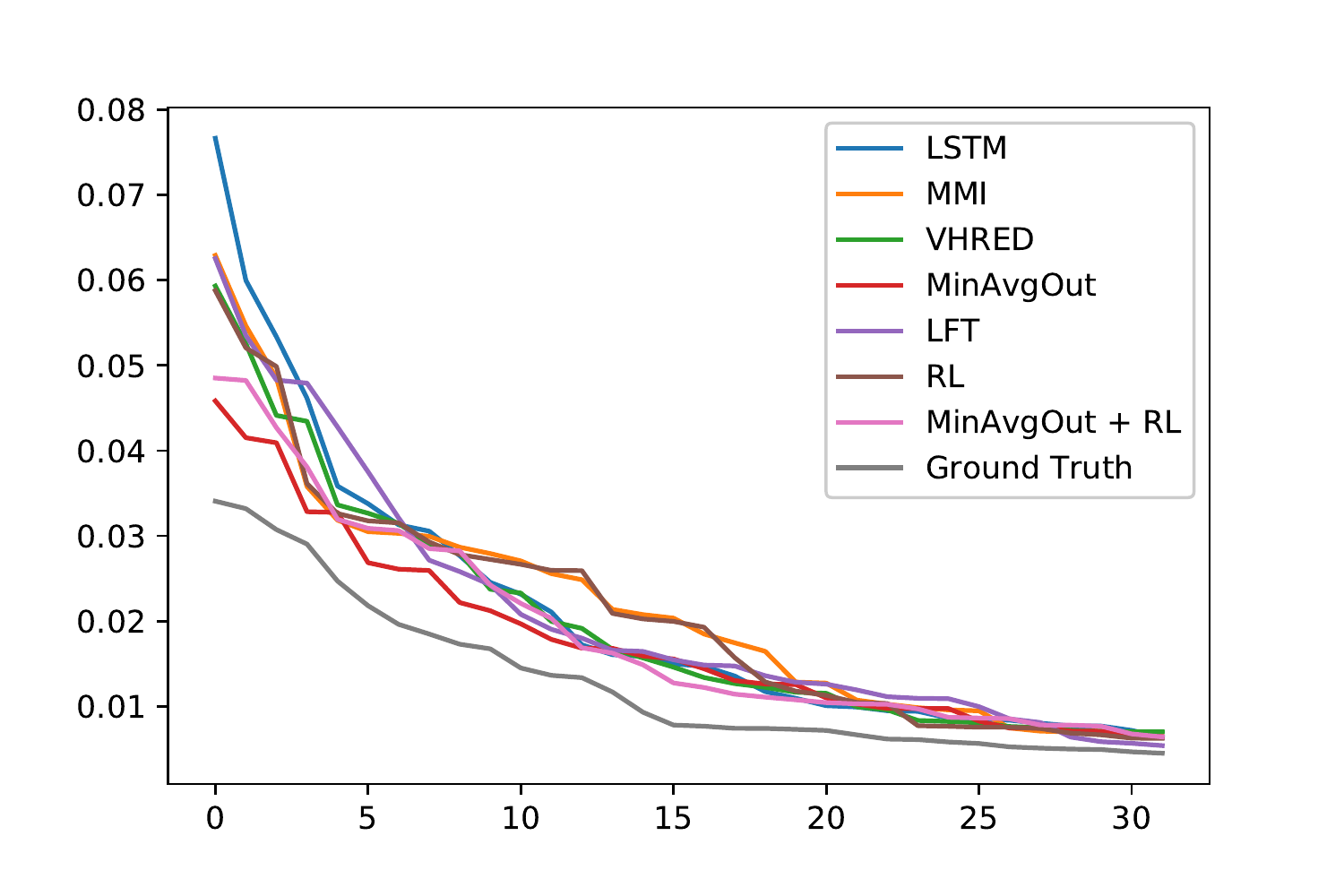}
        \caption[]%
        {{\small Unigram}}
        \label{fig:diversity-uni}
    \end{subfigure}
    \begin{subfigure}[b]{0.224\textwidth}   
        \centering 
        \includegraphics[width=\textwidth]{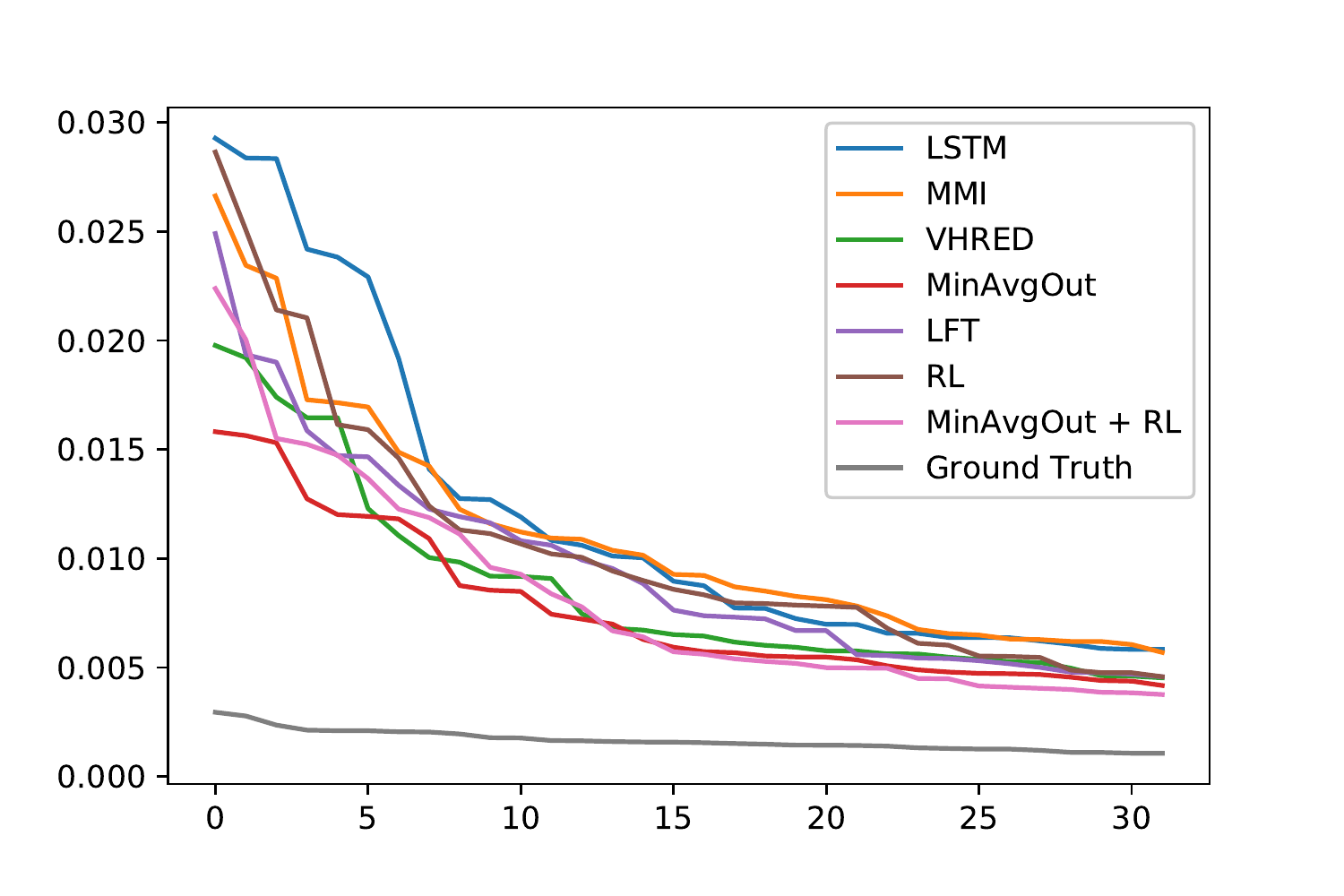}
        \caption[]%
        {{\small Bigram}}
        \label{fig:diversity-bi}
    \end{subfigure}
    \quad
    \begin{subfigure}[b]{0.224\textwidth}   
        \centering 
        \includegraphics[width=\textwidth]{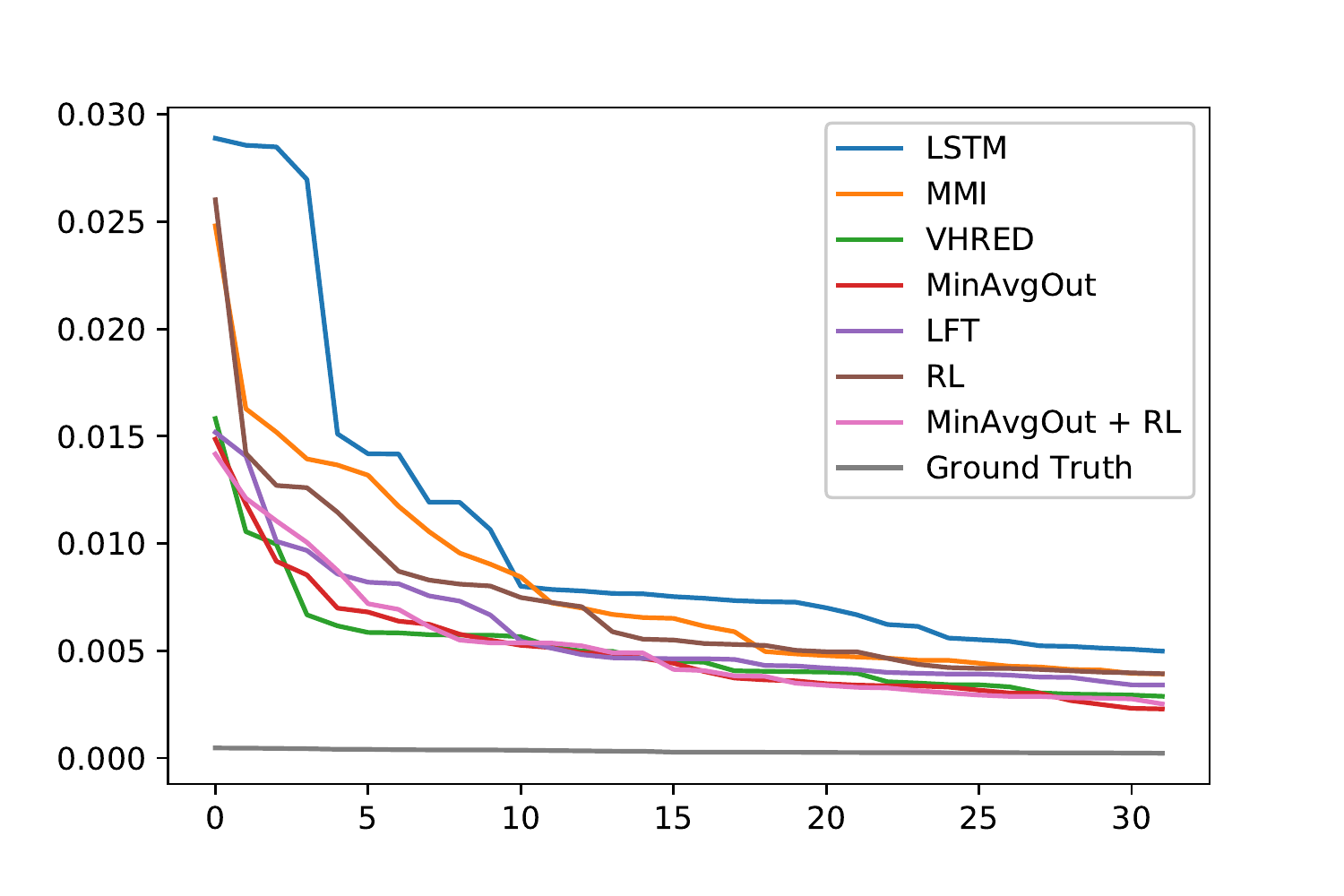}
        \caption[]%
        {{\small Trigram}}
        \label{fig:diversity-tri}
    \end{subfigure}
    \caption[]
    {\small Diversity-32 graphs of all models. Curves with lower AUC correspond to more diverse models.} 
    \label{fig:diversity-32}
\end{figure}

\subsection{Human Evaluation}
Although we proposed the effective AvgOut metric, we did find that the model sometimes still cheats to gain higher automatic diversity score. For example, as can be seen in the selected output examples (Section~\ref{sec:results and analysis}), the model tends to generate words with typo since these are rarer tokens as compared to their correct counterparts. This is unavoidable for noisy datasets like Ubuntu. Thus, without human evaluation, we can never be sure if our models are good or they only look good because our metrics are exploited.\footnote{This is also true for Ubuntu: the human studies and F-1 automatic evaluation are complementary to each other because MTurk annotators are better at judging how coherent and informative a response is, while F1s concentrate more on the technical terms (i.e., whether important activities or entities are present in the response).}

We thus also conducted human studies on Amazon MTurk to evaluate the generated responses with pairwise comparison for dialogue quality. We compare our models with an advanced decoding algorithm \textsc{MMI}~\cite{li2015diversity} and two models, namely \textsc{LSTM}~\cite{sordoni2015neural} and \textsc{VHRED}~\cite{serban2017hierarchical}, both with additive attention. 
To our best knowledge, LSTM and VHRED were the primary models with which F1's were reported on the Ubuntu dataset.
Following~\citeauthor{baheti2018generating} (\citeyear{baheti2018generating}), we employ two criteria: Plausibility and Content Richness. The first criterion measures whether the response is plausible given the context, while the second gauges whether the response is diverse and informative. The utterances were randomly shuffled to anonymize model identity. We only allowed annotators located in the US-located with at least an approval rate of $98\%$ and $10,000$ approved HITs. We collected $100$ annotations in total after rejecting those completed by people who assign exactly the same score to all model responses.
Since we evaluated $7$ models, we collected $700$ annotations in total, which came from a diverse pool of annotators.

\subsection{Training Details}
For each of the three models, the hidden size of the encoder is $256$, while the decoder hidden size is $512$. For \textsc{MinAvgOut}, the coefficient of the regularization loss term $\alpha$ is $100.0$; For \textsc{LFT}, during inference we feed a score of $0.015$ since it achieves a good balance between response coherence and diversity. For \textsc{RL}, the coefficient of the \textsc{RL} term $\beta$ is $100.0$. For the hybrid model \textsc{MinAvgOut} + \textsc{RL}, $\alpha$ and $\beta$ share a coefficient of $50.0$.

\begin{table}[t]
\small
  \centering
    \begin{tabular}{r|cc}
    \toprule
           & \multicolumn{1}{l}{Activity F1} & \multicolumn{1}{l}{Entity F1} \\
    \midrule
    LSTM & 1.18 & 0.87 \\
    LSTM (\textit{attn}) & 3.30 & 1.81 \\
    HRED & 4.34 & 2.22 \\
    VHRED & 4.63 & 2.53 \\
    MMI & 5.17 & 3.11 \\
    VHRED (\textit{attn}) & 5.94 & 3.52 \\
    Reranking-RL & 5.67 & 3.73 \\
    \hline
    MinAvgOut & 6.83 & 4.77 \\
    LFT & 7.55 & 5.05 \\
    RL & 5.80 & 3.62 \\
    MinAvgOut + RL & \textbf{8.05} & \textbf{5.47} \\
    \bottomrule
    \end{tabular}%
  \caption{Automatic Evaluation Activity/Entity F1 results for baselines and our 3 models (\textit{attn} means "with attention"). \textsc{LSTM}, \textsc{HRED} and \textsc{VHRED} are reported in~\citeauthor{serban2017multiresolution} (\citeyear{serban2017multiresolution}), \textsc{VHRED} (\textit{attn}) and Reranking-RL in~\citeauthor{niu2018adversarial} (\citeyear{niu2018adversarial}), and the rest are produced by our work. All our four models have statistically significantly higher F1 values ($p < 0.001$) against \textsc{VHRED} (\textit{attn}) and \textsc{MMI}.
  }
  \label{tab:f1}%
\end{table}%

\begin{table*}[t]
\small
  \centering
    \begin{tabular}{r|ccccc|cc}
    \toprule
          & 
          \multicolumn{1}{l}{\textit{i}AUC-s} &
          \multicolumn{1}{l}{\textit{i}AUC-1} &
          \multicolumn{1}{l}{\textit{i}AUC-2} &
          \multicolumn{1}{l}{\textit{i}AUC-3} &
          \multicolumn{1}{l}{\textit{i}AUC-avg} &
          \multicolumn{1}{l}{Distinct-1} & \multicolumn{1}{l}{Distinct-2} \\
    \midrule
    LSTM (\textit{attn}) & 0.545 & 0.310 & 0.614 & 0.665 & 0.5335 & 0.0074 & 0.0298 \\
    MMI & 0.794 & 0.320 & 0.645 & 0.746 & 0.6263 & 0.0076 & 0.0312 \\
    VHRED (\textit{attn}) & 0.832 & 0.364 & 0.725 & 0.836 & 0.6893 & 0.0076 & 0.0474\\
    \midrule
    MinAvgOut & 0.902 & \textbf{0.428} & \textbf{0.754} & \textbf{0.838} & \textbf{0.7305} & \textbf{0.0082} & \textbf{0.0507} \\
    LFT & \textbf{0.928} & 0.328 & 0.693 & 0.812 & 0.6903 & 0.0056 & 0.0301 \\
    RL & 0.848 & 0.328 & 0.662 & 0.769 & 0.6518 & 0.0074 & 0.0354 \\
    MinAvgOut+RL & 0.916 & 0.396 & 0.736 & 0.832 & 0.7200 & 0.0074 & 0.0442 \\
    \bottomrule
    \end{tabular}%
  \caption{Automatic Evaluation results for the baselines and our proposed models ("\textit{i}AUC" means "inverted AUC", or "1 - AUC"; "\textit{attn}" means "with attention"; "\textit{s}", "\textit{$1$}", "\textit{$2$}" and "\textit{$3$}" correspond to "sentence-level", "unigram", "bigram" and "trigram", respectively; \textit{i}AUC-avg is the average of all the other AUC columns). Best results are boldfaced. We do not calculate $p$-value because it does not apply to corpus-level metrics.
  }
  \label{tab:auto-eval-div}%
\end{table*}%

\begin{table}[t]
\small
  \centering
    \begin{tabular}{r|cccc}
    \toprule
          & \multicolumn{1}{l}{Plaus.} 
          & \multicolumn{1}{l}{Rich} 
          & \multicolumn{1}{l}{Avg.}
          & \multicolumn{1}{l}{SclDiff.} \\
    \midrule
    LSTM (\textit{attn}) & 3.46     & 2.62 & 3.04 & 0.28 \\
    MMI & 3.57 & 2.92 & 3.25 & 0.20 \\
    VHRED (\textit{attn}) & 3.54     & 3.12 & 3.33  & 0.13\\
    \hline
    MinAvgOut &  3.62    &  3.34  & 3.48 & \textbf{0.08} \\
    LFT   & \textbf{3.83}    & \textbf{3.47} & \textbf{3.65} & 0.10\\
    RL    & 3.61   & 2.88 & 3.25 & 0.22\\
    MinAvgOut+RL & 3.67 & 3.23 & 3.45 & 0.13\\
    \bottomrule
    \end{tabular}%
  \caption{Human Evaluation results for all the models we produce on Plausibility, Richness, average of the two, and scaled difference (difference between them divided by their average). Best Results are boldfaced. Note that for the last column, lower is better since we want balance between Plausibility and Content Richness. All results are pair-wise statistically significantly different with $p < 0.05$, except between \textsc{MinAvgOut} and \textsc{RL} on Plausibility, and between \textsc{MinAvgOut} and \textsc{MinAvgOut}+\textsc{RL} on Average.
  }
  \label{tab:human-eval}%
\end{table}%

\section{Results and Analysis}
\label{sec:results and analysis}
\subsection{Automatic Evaluation Results}
We employ several complementary metrics to capture different aspects of the model. The F1 results are shown in Table~\ref{tab:f1}.\footnote{Note that the F1 scores for this task are overall low because the conversations in the Ubuntu dataset are all open-ended. This is unlike tasks such as Question Answering where there is usually a correct response.} Among all single models, \textsc{LFT} performs the best, followed by \textsc{MinAvgOut}. \textsc{RL} is also comparable with previous state-of-the-art models \textsc{VHRED (\textit{attn})} and \textsc{Reranking-RL}.
We think that this is because LFT exerts no force in pulling the model predictions away from the ground-truth tokens, but rather just makes itself aware of how dull each response is. Consequently, its responses appear more relevant than the other two approaches.
Moreover, the hybrid model (last row) outperforms all other models by a large margin. One might expect that minimizing \textsc{AVGOUT} causes the models to move further away from the ground-truth tokens, so that it will hurt relevance. However, our F1 results show that as the responses become more diverse, they are more likely to include information more related and specific to the input contexts, which actually makes the model gain on both diversity and relevance. This will be further confirmed by the output examples in Table~\ref{tab:output}.

\begin{table*}[ht]
  \centering
    \begin{tabular}{r|p{13.0cm}}
        \toprule
        Context-X & do you think almoxarife 's network d/l idea is a good next step ? \\
        \hdashline
        Context-Y & the error message is from partman so its related to the disk partitions \\
        \midrule
        Seq2seq & what do you mean ? \\
        \midrule
        VRHED (attn) & it 's not showing up in fdisk -l \\
        \midrule
        MMI & it 's not a disk , it 's a windows machine \\
        \midrule
        MinAvgOut & I have a bunch of stuff in there , but I don't know how to do that \\
        \midrule
        LFT & it says " no such file or directory " \\
        \midrule
        RL & thanks for the tip \\
        \midrule
        MinAvgOut + RL & I did that , but it says that it is not mounted \\
        \midrule
        \midrule
        Context-X & im using the ubuntu default one  and depending on which I install it should set the home why run sudo update-alternatives -- config java then ?  that should set the JAVA\_HOME \\
        \hdashline
        Context-Y & env | grep -i java \\
        \midrule
        Seq2seq & what do you mean ? \\
        \midrule
        VRHED (attn) & i dont know how to do that \\
        \midrule
        MMI & thanks \\
        \midrule
        MinAvgOut & it says java is already the newest version \\
        \midrule
        LFT & it says that the java version is not installed \\
        \midrule
        RL & it says " java 1.4 . 2 " \\
        \midrule
        MinAvgOut + RL & ok , so what is the output of : sudo update-java-alternatives -l | grep java \\
        \bottomrule

    \end{tabular}%
  \caption{Selected output examples from all models. Context-X and -Y are given as model inputs during inference.
  }
  \label{tab:output}%
\end{table*}%

We also present \textsc{Diversity-32} graphs (Figure~\ref{fig:diversity-32}) and report \textsc{Diversity-AUC} as well as \textsc{Distinct-1} and \textsc{-2} for each model (Table~\ref{tab:auto-eval-div}). We can see that all our models have significantly better sentence-level diversity than \textsc{VHRED}, let alone \textsc{LSTM}. For unigram diversity, they are also better than \textsc{LSTM}, though hard to distinguish from \textsc{VHRED}. Both bigram and trigram graphs reveal that all models are more diverse than \textsc{LSTM}, except that \textsc{RL} shows lower diversity than the other models, which agree with our F1 results. Note that since our models are only trained based on unigram output distributions, the bigram and trigram diversities are still far away from that of the ground-truth, which points to future direction. 
That said, the table does show that encouraging unigram diversity can already have positive influence on higher grams as well. Also note that the hybrid model (last row) does not achieve the best result in terms of diversity. We hypothesize that this is because \textsc{RL}, which is usually harder to optimize than \textsc{ML} losses, faces exacerbated issues when combined with a strong \textsc{MinAvgOut} loss, which tries to pull the model output distribution away from the token distribution in the training corpus.

Neither \textsc{Distinct-1} nor \textsc{-2} correlates well with our observation and evaluation of diversity and relevance. We reason that this is because these metrics only capture how many distinct tokens are used rather than each token's frequency, which is easier to be exploited because whether each token is used unnecessarily often (a strong sign of dullness) is not reflected in this measure.

\subsection{Human Evaluation Results}
As mentioned in experimental setup, we conducted human evaluations on our models for both Plausibility and Content Richness, as well as calculating their average (to show overall score) and their difference (to show balance between the two criteria) (Table~\ref{tab:human-eval}). We can see from the table that all our models are statistically significantly better than the baseline models on both Plausibility and Content Richness, except that \textsc{RL} is slightly weaker on Content Richness, which agrees with the trend in automatic evaluations. Although \textsc{MinAvgOut}+\textsc{RL} model only ranks the second on average score (statistically equivalent to \textsc{MinAvgOut}) in human evaluation, it achieves a good balance, and it also ranks the second in automatic diversity and the first in F1 values. We thus consider it to be our best model.

\subsection{Selected Output Examples}
We present two selected examples of generated responses from the investigated models (Table~\ref{tab:output}). We can see that all our models learn to attend well to the context, generating coherent and informative responses.

\section{Related Work}
\label{sec:related work}
\subsection{Measurements of Response Diversity}
Multiple metrics and approaches have been proposed to measure dialogue diversity. Some focus more on how similar the responses are to the ground-truth sequences, such as Word Error Rate~\cite{serban2016building} and BLEU~\cite{galley2015deltableu}, while the others explicitly have diversity in mind when being created, such as \textsc{Distinct-1} and \textsc{-2}~\cite{li2015diversity}. The key difference between AvgOut and the previous work is that first, our metric is dynamic with no feature-engineering; second, ours is versatile enough to be applied to both continuous distributions and discrete sequences, while theirs are only for discrete tokens; third, ours can be used for both sentence-level and corpus-level evaluation, while theirs are only meaningful as corpus-level metrics because they measure the extent of repetition across responses rather than for a single response.

\subsection{Diversity-Promoting Dialogue Models}
Researchers have different opinions on why dull responses are generated, which lead to various solutions. They can be roughly divided into four categories. The first category considers using conditional likelihood as a decoding objective the culprit~\cite{baheti2018generating,li2015diversity,li2017learning,shao2017generating}. They thus focus on improving the decoding algorithm during training. The second category traces the cause of the low-diversity problem back to the lack of model variability. They then adopt Variational Autoencoders and rely on sampling from a latent random variable as an additional prior to the decoder~\cite{serban2017hierarchical,zhao2017learning,gao2019jointly}. The third category thinks that the issue is a lack of universal background knowledge and common sense beyond the input context. They consequently aim to integrate prior knowledge into the generation process~\cite{raghu2018hierarchical,liu2018knowledge,pei2018s2spmn,kryscinski2018improving}. 
The fourth category believes that the underlying model itself needs improvement. Some use hierarchical LSTM-RNN to encourage the model to capture high-level context~\cite{serban2016building}; some use more advanced attention mechanism such as multi-head attention~\cite{tao2018get}; and some use either more complicated architectures or models prone to degeneracies, such as Generative Adversarial Networks~\cite{li2017adversarial}, Deep Reinforcement Learning~\cite{li2016deep} and Mixture Models~\cite{shen2019mixture}. Our \textsc{RL} model has the same architecture as the Reinforcement Learning model, except with different rewards.
~\citeauthor{jiang2018sequence} (\citeyear{jiang2018sequence}) consider the reason for dull responses as the model's over-confidence. They then propose to add to the loss function a regularization term to maximize the entropy of the output probability distribution. Interestingly, they only proposed this simple approach rather than actually implementing it. Our \textsc{MinAvgOut} approach is related to their idea.
Our approach is also related to posterior regularization~\cite{mann2008generalized,ganchev2010posterior,zhu2014bayesian}, but our work is neural-based.

\section{Conclusion}
\label{sec:conclusion}
We proposed a novel measure \textsc{AvgOut} to dynamically evaluate how diverse a model or a response is based on the models' own parameters, which themselves evolve during training. We then leveraged this effective measure to train three models, plus a hybrid model, to eliminate dull responses for dialogue generation tasks. In addition, we designed novel automatic metrics to evaluate the trained models on diversity, in addition to the ones from previous work. Both automatic and human evaluations consolidated that our models are able to generate more diverse and relevant responses, even when compared with state-of-the-art approaches. For future work, we plan to apply these models to different generative tasks where diversity is desired.

\section*{Acknowledgments}
We thank the reviewers for their helpful comments. This work was supported by NSF-CAREER Award \#1846185, ONR \#N00014-18-1-2871, and awards from Google, Facebook, Salesforce (views are not of the funding agency).

\bibliography{references}

\begin{thebibliography}{}

\bibitem[\protect\citeauthoryear{Bahdanau, Cho, and
  Bengio}{2015}]{Bahdanau2015}
Bahdanau, D.; Cho, K.; and Bengio, Y.
\newblock 2015.
\newblock Neural machine translation by jointly learning to align and
  translate.
\newblock In {\em Proceedings of ICLR}.

\bibitem[\protect\citeauthoryear{Baheti \bgroup et al\mbox.\egroup
  }{2018}]{baheti2018generating}
Baheti, A.; Ritter, A.; Li, J.; and Dolan, B.
\newblock 2018.
\newblock Generating more interesting responses in neural conversation models
  with distributional constraints.
\newblock In {\em Proceedings of EMNLP}.

\bibitem[\protect\citeauthoryear{Galley \bgroup et al\mbox.\egroup
  }{2015}]{galley2015deltableu}
Galley, M.; Brockett, C.; Sordoni, A.; Ji, Y.; Auli, M.; Quirk, C.; Mitchell,
  M.; Gao, J.; and Dolan, B.
\newblock 2015.
\newblock deltableu: A discriminative metric for generation tasks with
  intrinsically diverse targets.
\newblock In {\em Proceedings of ACL}.

\bibitem[\protect\citeauthoryear{Ganchev \bgroup et al\mbox.\egroup
  }{2010}]{ganchev2010posterior}
Ganchev, K.; Gillenwater, J.; Taskar, B.; et~al.
\newblock 2010.
\newblock Posterior regularization for structured latent variable models.
\newblock {\em JMLR} 11(Jul):2001--2049.

\bibitem[\protect\citeauthoryear{Gao \bgroup et al\mbox.\egroup
  }{2019}]{gao2019jointly}
Gao, X.; Lee, S.; Zhang, Y.; Brockett, C.; Galley, M.; Gao, J.; and Dolan, B.
\newblock 2019.
\newblock Jointly optimizing diversity and relevance in neural response
  generation.
\newblock In {\em NAACL}.

\bibitem[\protect\citeauthoryear{Hochreiter and
  Schmidhuber}{1997}]{hochreiter1997long}
Hochreiter, S., and Schmidhuber, J.
\newblock 1997.
\newblock Long short-term memory.
\newblock {\em Neural computation} 9(8):1735--1780.

\bibitem[\protect\citeauthoryear{Jiang and de Rijke}{2018}]{jiang2018sequence}
Jiang, S., and de~Rijke, M.
\newblock 2018.
\newblock Why are sequence-to-sequence models so dull? understanding the
  low-diversity problem of chatbots.
\newblock In {\em Proceedings of EMNLP}.

\bibitem[\protect\citeauthoryear{Kikuchi \bgroup et al\mbox.\egroup
  }{2016}]{DBLP:journals/corr/KikuchiNSTO16}
Kikuchi, Y.; Neubig, G.; Sasano, R.; Takamura, H.; and Okumura, M.
\newblock 2016.
\newblock Controlling output length in neural encoder-decoders.
\newblock In {\em EMNLP},  1328–--1338.

\bibitem[\protect\citeauthoryear{Kry{\'s}ci{\'n}ski \bgroup et al\mbox.\egroup
  }{2018}]{kryscinski2018improving}
Kry{\'s}ci{\'n}ski, W.; Paulus, R.; Xiong, C.; and Socher, R.
\newblock 2018.
\newblock Improving abstraction in text summarization.
\newblock In {\em ACL}.

\bibitem[\protect\citeauthoryear{Li \bgroup et al\mbox.\egroup
  }{2016a}]{li2015diversity}
Li, J.; Galley, M.; Brockett, C.; Gao, J.; and Dolan, B.
\newblock 2016a.
\newblock A diversity-promoting objective function for neural conversation
  models.
\newblock In {\em Proceedings of NAACL}.

\bibitem[\protect\citeauthoryear{Li \bgroup et al\mbox.\egroup
  }{2016b}]{li2016deep}
Li, J.; Monroe, W.; Ritter, A.; Galley, M.; Gao, J.; and Jurafsky, D.
\newblock 2016b.
\newblock Deep reinforcement learning for dialogue generation.
\newblock In {\em Proceedings of ACL}.

\bibitem[\protect\citeauthoryear{Li \bgroup et al\mbox.\egroup
  }{2017}]{li2017adversarial}
Li, J.; Monroe, W.; Shi, T.; Jean, S.; Ritter, A.; and Jurafsky, D.
\newblock 2017.
\newblock Adversarial learning for neural dialogue generation.
\newblock In {\em Proceedings of ACL}.

\bibitem[\protect\citeauthoryear{Li, Monroe, and
  Jurafsky}{2017}]{li2017learning}
Li, J.; Monroe, W.; and Jurafsky, D.
\newblock 2017.
\newblock Learning to decode for future success.
\newblock {\em arXiv preprint arXiv:1701.06549}.

\bibitem[\protect\citeauthoryear{Liu \bgroup et al\mbox.\egroup
  }{2016}]{liu2016not}
Liu, C.-W.; Lowe, R.; Serban, I.~V.; Noseworthy, M.; Charlin, L.; and Pineau,
  J.
\newblock 2016.
\newblock How {NOT} to evaluate your dialogue system: An empirical study of
  unsupervised evaluation metrics for dialogue response generation.
\newblock In {\em EMNLP}.

\bibitem[\protect\citeauthoryear{Liu \bgroup et al\mbox.\egroup
  }{2018}]{liu2018knowledge}
Liu, S.; Chen, H.; Ren, Z.; Feng, Y.; Liu, Q.; and Yin, D.
\newblock 2018.
\newblock Knowledge diffusion for neural dialogue generation.
\newblock In {\em Proceedings of ACL}, volume~1,  1489--1498.

\bibitem[\protect\citeauthoryear{Lowe \bgroup et al\mbox.\egroup
  }{2015}]{lowe2015ubuntu}
Lowe, R.; Pow, N.; Serban, I.; and Pineau, J.
\newblock 2015.
\newblock The ubuntu dialogue corpus: A large dataset for research in
  unstructured multi-turn dialogue systems.
\newblock In {\em SIGDIAL}.

\bibitem[\protect\citeauthoryear{Lowe \bgroup et al\mbox.\egroup
  }{2017}]{lowe2017towards}
Lowe, R.; Noseworthy, M.; Serban, I.~V.; Angelard-Gontier, N.; Bengio, Y.; and
  Pineau, J.
\newblock 2017.
\newblock Towards an automatic turing test: Learning to evaluate dialogue
  responses.
\newblock In {\em ACL}.

\bibitem[\protect\citeauthoryear{Mann and McCallum}{2008}]{mann2008generalized}
Mann, G.~S., and McCallum, A.
\newblock 2008.
\newblock Generalized expectation criteria for semi-supervised learning of
  conditional random fields.
\newblock {\em ACL-08: HLT}  870.

\bibitem[\protect\citeauthoryear{Niu and Bansal}{2018a}]{niu2018adversarial}
Niu, T., and Bansal, M.
\newblock 2018a.
\newblock Adversarial over-sensitivity and over-stability strategies for
  dialogue models.
\newblock In {\em CoNLL}.

\bibitem[\protect\citeauthoryear{Niu and Bansal}{2018b}]{TACL1424}
Niu, T., and Bansal, M.
\newblock 2018b.
\newblock Polite dialogue generation without parallel data.
\newblock {\em TACL} 6:373--389.

\bibitem[\protect\citeauthoryear{Papineni \bgroup et al\mbox.\egroup
  }{2002}]{Papineni:2002:BMA:1073083.1073135}
Papineni, K.; Roukos, S.; Ward, T.; and Zhu, W.-J.
\newblock 2002.
\newblock {BLEU}: A method for automatic evaluation of machine translation.
\newblock In {\em Proceedings of ACL},  311--318.

\bibitem[\protect\citeauthoryear{Paulus, Xiong, and
  Socher}{2018}]{DBLP:journals/corr/PaulusXS17}
Paulus, R.; Xiong, C.; and Socher, R.
\newblock 2018.
\newblock A deep reinforced model for abstractive summarization.
\newblock In {\em ICLR}.

\bibitem[\protect\citeauthoryear{Pei and Li}{2018}]{pei2018s2spmn}
Pei, J., and Li, C.
\newblock 2018.
\newblock S2spmn: a simple and effective framework for response generation with
  relevant information.
\newblock In {\em Proceedings of EMNLP},  745--750.

\bibitem[\protect\citeauthoryear{Raghu, Gupta, and
  others}{2019}]{raghu2018hierarchical}
Raghu, D.; Gupta, N.; et~al.
\newblock 2019.
\newblock Hierarchical pointer memory network for task oriented dialogue.
\newblock In {\em NAACL}.

\bibitem[\protect\citeauthoryear{Ranzato \bgroup et al\mbox.\egroup
  }{2016}]{ranzato2015sequence}
Ranzato, M.; Chopra, S.; Auli, M.; and Zaremba, W.
\newblock 2016.
\newblock Sequence level training with recurrent neural networks.
\newblock In {\em Proceedings of ICLR}.

\bibitem[\protect\citeauthoryear{Serban \bgroup et al\mbox.\egroup
  }{2016}]{serban2016building}
Serban, I.~V.; Sordoni, A.; Bengio, Y.; Courville, A.; and Pineau, J.
\newblock 2016.
\newblock Building end-to-end dialogue systems using generative hierarchical
  neural network models.
\newblock In {\em AAAI}.

\bibitem[\protect\citeauthoryear{Serban \bgroup et al\mbox.\egroup
  }{2017a}]{serban2017multiresolution}
Serban, I.~V.; Klinger, T.; Tesauro, G.; Talamadupula, K.; Zhou, B.; Bengio,
  Y.; and Courville, A.~C.
\newblock 2017a.
\newblock Multiresolution recurrent neural networks: An application to dialogue
  response generation.
\newblock In {\em Proceedings of AAAI},  3288--3294.

\bibitem[\protect\citeauthoryear{Serban \bgroup et al\mbox.\egroup
  }{2017b}]{serban2017hierarchical}
Serban, I.~V.; Sordoni, A.; Lowe, R.; Charlin, L.; Pineau, J.; Courville, A.;
  and Bengio, Y.
\newblock 2017b.
\newblock A hierarchical latent variable encoder-decoder model for generating
  dialogues.
\newblock In {\em AAAI}.

\bibitem[\protect\citeauthoryear{Shao \bgroup et al\mbox.\egroup
  }{2017}]{shao2017generating}
Shao, L.; Gouws, S.; Britz, D.; Goldie, A.; Strope, B.; and Kurzweil, R.
\newblock 2017.
\newblock Generating high-quality and informative conversation responses with
  sequence-to-sequence models.
\newblock In {\em Proceedings of EMNLP}.

\bibitem[\protect\citeauthoryear{Shen \bgroup et al\mbox.\egroup
  }{2019}]{shen2019mixture}
Shen, T.; Ott, M.; Auli, M.; and Ranzato, M.
\newblock 2019.
\newblock Mixture models for diverse machine translation: Tricks of the trade.
\newblock In {\em Proceedings of ICML}.

\bibitem[\protect\citeauthoryear{Sordoni \bgroup et al\mbox.\egroup
  }{2015}]{sordoni2015neural}
Sordoni, A.; Galley, M.; Auli, M.; Brockett, C.; Ji, Y.; Mitchell, M.; Nie,
  J.-Y.; Gao, J.; and Dolan, B.
\newblock 2015.
\newblock A neural network approach to context-sensitive generation of
  conversational responses.
\newblock In {\em Proceedings of NAACL}.

\bibitem[\protect\citeauthoryear{Sutton \bgroup et al\mbox.\egroup
  }{2000}]{Sutton00policygradient}
Sutton, R.~S.; Mcallester, D.; Singh, S.; and Mansour, Y.
\newblock 2000.
\newblock Policy gradient methods for reinforcement learning with function
  approximation.
\newblock In {\em NeurIPS}.

\bibitem[\protect\citeauthoryear{Tao \bgroup et al\mbox.\egroup
  }{2018}]{tao2018get}
Tao, C.; Gao, S.; Shang, M.; Wu, W.; Zhao, D.; and Yan, R.
\newblock 2018.
\newblock Get the point of my utterance! learning towards effective responses
  with multi-head attention mechanism.
\newblock In {\em Proceedings of IJCAI},  4418--4424.

\bibitem[\protect\citeauthoryear{Vinyals and Le}{2015}]{vinyals2015neural}
Vinyals, O., and Le, Q.
\newblock 2015.
\newblock A neural conversational model.
\newblock In {\em Proceedings of ICML}.

\bibitem[\protect\citeauthoryear{Zhao, Zhao, and
  Eskenazi}{2017}]{zhao2017learning}
Zhao, T.; Zhao, R.; and Eskenazi, M.
\newblock 2017.
\newblock Learning discourse-level diversity for neural dialog models using
  conditional variational autoencoders.
\newblock In {\em Proceedings of ACL}.

\bibitem[\protect\citeauthoryear{Zhu, Chen, and Xing}{2014}]{zhu2014bayesian}
Zhu, J.; Chen, N.; and Xing, E.~P.
\newblock 2014.
\newblock Bayesian inference with posterior regularization and applications to
  infinite latent svms.
\newblock {\em JMLR} 15(1):1799--1847.

\end{thebibliography}
\bibliographystyle{aaai}

\end{document}